# An Application Of Backpropagation Artificial Neural Network Method for Measuring The Severity of Osteoarthritis

Dian Pratiwi[1], Diaz D. Santika[2], and Bens Pardamean[3]

[1]Information Engineering Department, Trisakti University, Jakarta, Indonesia
[2]Information Technology Department, Bina Nusantara University, Jakarta, Indonesia
[3]Information Technology Department, Bina Nusantara University, Jakarta, Indonesia

The examination of Osteoarthritis disease through X-ray by rheumatology can be classified into four grade of severity. This paper discusses about the application of artificial neural network backpropagation method for measuring the severity of the disease, where the observed X-ray range from wrist to fingers.

The main procedures of system in this paper is divided into three, which are image processing, feature extraction, and artificial neural network process. First, an X-ray image digital (200x150 pixels and greyscale) will be thresholded, then extracted features based on probabilistic values of the color intensity of seven bit quantization result, and statistical textures.

That feature values then will be normalizing to interval [0.1, 0.9], and then the result would be processing on backpropagation artificial neural network system as input to determine the severity of disease from an X-ray had input before it. From testing with learning rate 0.3, momentum 0.4, hidden units five pieces and about 132 feature vectors, this system had had a level of accuracy of 100% for learning data, 80% for learning and non-learning data, and 66.6% for non-learning data

## I. INTRODUCTION

The disease know as osteoarthritis or calcification generally occurs in areas around joints such as cartilage and joints the fingers, knees, and spine that causes symptoms of fatigue, inflammation, pain and swelling that accompanied reddish color. These symptoms are used as an initial analysis by rheumatologist. However, in ensuring conditions of osteoarthritis not simply on the basis of the symptoms that arise. Because of many cases showed that advanced stage osteoarthritis are not or only slightly showing symptoms, and vice versa. Therefore, further tests such as X-rays, Magnetic Resonance Imaging (MRI), or Computed Tomography Imaging (CT Scan) is recommended to do in order to give more accurate results.

In this paper, we discuss a classification technique based on Artificial Neural Network (ANN) is applied by the author in order to help predict or measure the severity of the osteoarthritis disease through X-ray image from the fourth grade. The fourth grade have a different description of the osteoarthritis condition .and only can be diagnosed by rheumatologist, radiologist, or orthopedist through a series of medical examinations.

Some images are used by orthopedist or rheumatologist in determining the severity of osteoarthritis disease from X-ray photograph is to see whether or not cysts, swelling, formation of new bone that is not flat and spiky (osteophyte), subchondral sclerosis, and reducing cartilage mass [1]. The description of these condition then made it as a reference in assisting rheumatologist ensure osteoarthritis disease from the X-ray image set.

Every characteristic and feature of osteoarthritis that appear on X-ray image will be processed and extracted based on the brightness (intensity) and texture, which then will be processed to generate predictions in software-based artificial neural network backpropagation method.

*1. Goal*

The goal in this paper, ie :
  a) Obtaining a correct prediction results in a measure the severity of osteoarthritis disease of each X-ray image
  b) Generate one way to classify the level of osteoarthritis disease is by applying the method of backpropagation artificial neural network.

*2. Benefits*

The benefits in this paper are :
  a) Helps rheumatologist in ensuring the severity of osteoarthritis from patient's condition, so that treatment can be given with no undertreatment or overtreatment
  b) Can be used as a basis of preliminary design or develop a system of measuring the severity of other diseases, which can be applied in subsequent studies.

## II. SAMPLE AND PROCEDURE

*A. Sampel Data*

The data used in this study is a set of X-ray images derived from patients with osteoarthritis with a variety of conditions. This condition can be divided into four grades of severity with a different picture, namely [1] :
  a. Grade 1 (Doubtful) :
   • Distal interphalangeal joints : Normal joint except for one minimal osteophyte.
   • Proximal interphalangeal joints : Minimal osteophytosis at one point and possible cyst
   • First carpometacarpal joint : Minimal osteophytosis and possible cyst formation
  b. Grade 2 (Minimal) :
   • Distal interphalangeal joints : Definite osteophytes at two points with minimal subchondral sclerosis and doubtful subchondral cysts, but good joint space and no deformity

- Proximal interphalangeal joints : Definite osteophytes at two points and possible narrowing of joint space at one point
- First carpometacarpal joint : Definite osteophytes and possible cysts

c. Grade 3 (Moderate) :
- Distal interphalangeal joints : Moderate osteophytes, some deformity of bone ends and narrowing of joint space
- Proximal interphalangeal joints : Moderate osteophytes at many points, deformity of bone ends
- First carpometacarpal joint : Moderate osteophytes, narrowing of joint space, and subchondral sclerosis and deformity of bone ends

d. Grade 4 (Severe) :
- Distal interphalangeal joints : Large osteophytes and deformity of bone ends with loss of joint space, sclerosis, and cysts
- Proximal interphalangeal joints : Large osteophytes, marked narrowing of joint space, subchondral sclerosis, and slight deformity
- First carpometacarpal joint : Large osteophytes, severe sclerosis, and narrowing of joint space

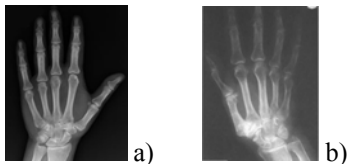

Fig. 1. X-ray image of hands. (a) X-ray image of non-osteoarthritis (b) X-ray image of osteoarthritis [2]

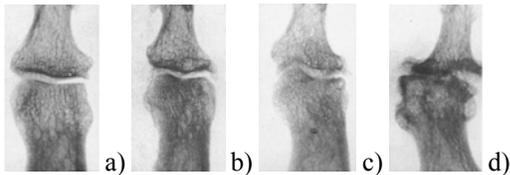

Fig. 2 [1]. X-ray image of Osteoarthritis in the various grades of The Distal Interphalangeal joints ; (a) Grade 1, (b) Grade2, (c) Grade 3, (d) Grade 4

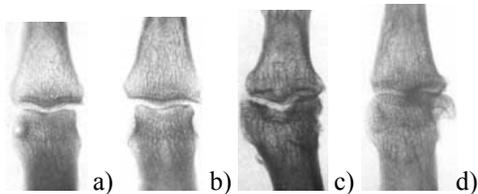

Fig. 3 [1]. X-ray image of Osteoarthritis in the various grades of The Proximal Interphalangeal joints ; (a) Grade 1, (b) Grade 2, (c) Grade 3, (d) Grade 4

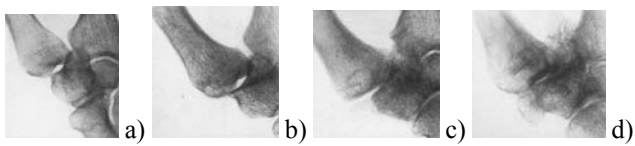

Fig. 4[1]. X-ray image of Osteoarthritis in the various grades of The First Carpometacarpal joint ; (a) Grade 1, (b) Grade 2, (c) Grade 3, (d) Grade 4

X-ray region were observed and ranged from wrist to fingers with a size of 200x150 pixels. The number of samples that will be used is 60 x-ray images.

*B. Procedure*

In this study, there is a set of processes undertaken to produce a prediction system and the percentage of measurement accuracy.

1. Image processing
   a. Convert RGB color to grayscale
   Convert color image of The X-ray done to change the color of RGB (24 bit) into grayscale (8 bit), so that the gray value at interval 0 to 255.
   b. Thresholding
   A process to classify the pixels within the limits of particular intensity, and separates the image between the foreground with the background [3]. In this study, the appropriate threshold value is taken through a process of trial and error for more or less suitable for all images [4].

2. Feature extraction

Feature extraction is the first step in conducting a classification and image interpretation. The process associated with the quantization characteristic of the image into the appropriate group of feature values. With the feature extraction, important information in the image can be captured and stored into a feature vector.

In this study, the extracted features are based on the intensity of color and texture.

a. Feature extraction based on color intensity

The intensity of color obtained from the simplification of the three components RGB into one component intensity with grey level value between 0 – 255. Distribution of intensity values of an image can be viewed via a histogram, which the probability of every grey level value can be taken using the formula :

$$h_i = \frac{n_i}{n} \quad (1)$$

Where $n_i$ is the number of pixels that have grey level value i (i = 0... L = 1), $L$ is the maximum interval of color, $n$ is the total pixels of image, and $h_i$ is the probability of $i$ grey level value.

Grey level values then can be grouped into several levels through the process of quantization. Quantization will divide the grayscale [0, L-1] into G levels are expressed as an integer.

$$G = 2^m \quad (2)$$

Where $G$ is the grey level, and $m$ is the positive integer or the number of bits

b. Feature extraction based on texture

Image feature extraction based on first-order texture can use statistical methods, namely by looking at the statistical distribution of grey level on the image histogram [5]. From the values in the histogram, can be calculated feature parameters include :

1) Variance ($\sigma^2$)

Shows the variation element of the histogram of an image

$$\sigma^2 = \sum_n (f_n - \mu)^2 p(f_n) \quad (3)$$

Where $f_n$ is the intensity value of grey level, $\mu$ is the average value of grey level intensity, $p(f_n)$ is the value of histogram ( probability of occurence of intensity in the image)

2) Skewness ($\alpha_3$)

Shows the level of relative inclination of an image histogram curve

$$\alpha_3 = \frac{1}{\sigma^3} \sum_n (f_n - \mu)^3 p(f_n) \quad (4)$$

3) Entropy (H)

Shows the shape of an irregular size image

$$H = -\sum_n p(f_n) \cdot {}^2\!\log p(f_n) \quad (5)$$

4) Relative smoothness (R) [6]

Shows the level of relative smoothness of an image form.

$$R = 1 - (1/1 + \sigma^2) \quad (6)$$

3. Normalization of Feature Vector

Normalization is a method to classify the range or interval of values that are different to the same scale and smaller. In this study, Min-Max Normalization is used as normalization method [7] .

$$D'(i) = \frac{D(i) - min(D)}{max(D) - min(D)} * (U - L) + L \quad (7)$$

Where $D$ is the natural data, $U$ and $L$ are the upper and lower normalization bound.

In this study, the use of normalization will be applied to classify the values of the different features into the range between 0.1 to 0.9.

4. Classification with Backpropagation Neural Network

After pre-processing step performed, the values of the resulting features will the be processed in The Backpropagation Neural Network. From this process will be obtained prediction of disease all the data which will then determine the percentage of success of this method.

III. RESULTS AND DISCUSSION

The tests in this study consists of image processing phase, feature vector extraction phase, and three network phases that is the network training phase, validation phase, and final testing phase.  In the training phase, x-ray image that is used as many as 36 pieces from 60% of the total data, with each level of severity of disease was nine images. While the validation phase, x-ray are used as much as 20 images, with each level of severity of disease was five images. And in the final testing phase, the x-ray that was used as many as 12 images.

A. Image Processing (Pre-Processing)

In this study, image processing phase is done first, ie the conversion of X-ray color from the RGB color into greyscale. After that, every X-ray then going through the process of thresholding to take a detailed framework of the bone area of hands (especially the joints) which is needed in diagnosing the osteoarthritis disease.

Threshold value that is used on the results obtained from the thresholding process of several X-ray images trial and error between 30 to 70.

From the test results, use the threshold value of 50 has made some parts of detail object framework of bone is reduced. This will cause the number of missing essential information, so that the results of the osteoarthritis level classification on the network will be less than the maximum possible. While the use of threshold of 30, the background is still not a whole successfully removed and this will cause the pixels will be considered as a foreground that allows the network will be difficult to recognize pattern correctly. So that we determined the threshold value in the thresholding process is equal to 40, because of the test results showed that these values are best used in a separate object area with a background and able to maintain the detail object.

B. Feature Vector Extraction

Feature extraction process that is conducted in this study is based on the categories of color and texture. Color features is taken through the calculation of the probability histogram (interval 0 to 255) which then go through the quantization will generate the values of the features as much as 128 features. While on the texture, the feature is taken through the calculating formula of the value of statistical variance, skewness, entropy, and relative smoothness. Thus, the overall number of features used is about 132 features.

The following is Table I that contains some texture feature values of eight training images (with a total of two X-ray images of each grade) are extracted.

TABLE I. TEXTURE FEATURE VALUE OF DIFFERENT TRAINING IMAGES

| No | Texture Features | | | |
|---|---|---|---|---|
| | Variance | Skewness | Entropy | Relative Smoothness |
| 1 | 2542.55 | -0.644 | 0.704 | 1 |
| 2 | 1550.28 | -0.968 | 0.819 | 0.999 |
| 3 | 3959.754 | -1.174 | 0.922 | 1 |
| 4 | 605.424 | 0.128 | 0.495 | 0.998 |
| 5 | 3498.11 | -1.19 | 0.885 | 1 |
| 6 | 3799.3 | -1.143 | 0.865 | 1 |
| 7 | 2246.89 | -1.196 | 0.987 | 1 |
| 8 | 997.393 | -0.248 | 0.592 | 0.999 |

C. Training Phase

At this phase, all feature vectors that have been generated from the feature extraction and normalization process will be trained repeatedly in the network to get the best weight value. These weights resulting from training that has reached 100% recognition rate, ie when the Mean Square Error (MSE) value was smaller then the value of tolerance. These weights then will be used in the validation and testing phase.

The configuration of neural network parameters which have given in the training process in this study is three layers (one input layer, one hidden layer, and one output layer), two units of output, binary sigmoid activation function, threshold 0.5, momentum 0.5, learning rate 0.3, tolerance 0.001. This study also used Nguyen Widrow method for the weight and bias initialization.

From the results of experiments on the training phase using 36 images (training data), the value of learning rate of 0.3, learning rate of 0.5, and five hidden units has resulted 100% of recognition percentage when the 20$^{th}$ epoch, where each weight then will be stored for use at validation and final testing phase.

*D. Validation Phase*

In this study, the validation phase will be used to determine how likely level of success and error recognition of this ANN based tool for measuring the severity of osteoarthritis's disease prior to testing. The network process consists of feed forward only step, which will provide a direct output of data in the form of predictive learning and non-learning are included.

TABLE II. GRADE OR LEVEL MEASUREMENT RESULTS OF OSTEOARTHRITIS'S DISEASE IN VALIDATION PHASE

| Measurement Results | | | | Rheumatologists Diagnosis |
|---|---|---|---|---|
| Grade 1 | Grade 2 | Grade 3 | Grade 4 | |
| 4 | 1 | 0 | 0 | Grade 1 |
| 0 | 3 | 1 | 1 | Grade 2 |
| 0 | 1 | 4 | 0 | Grade 3 |
| 0 | 0 | 0 | 5 | Grade 4 |

From Table II can be seen, that the four of 20 x-ray images which included there's still have an error prediction. This can be caused by the part of tested data is new data that have not been trained, so that there's still a possibility of error recognition. However, the percentage of success obtained by 80%, so it can be concluded that these tools are quite capable for use in predicting the disease severity of osteoarthritis in the process of final testing.

*E. Testing Phase*

Same as validation phase, testing phase also just run the feed forward of testing data using the weights from the previous training process.

TABLE III. GRADE OR LEVEL MEASUREMENT RESULTS OF OSTEOARTHRITIS'S DISEASE IN TESTING PHASE

| Measurement Results | | | | Rheumatologists Diagnosis |
|---|---|---|---|---|
| Grade 1 | Grade 2 | Grade 3 | Grade 4 | |
| 3 | 0 | 0 | 0 | Grade 1 |
| 0 | 2 | 1 | 0 | Grade 2 |
| 0 | 1 | 1 | 1 | Grade 3 |
| 0 | 0 | 1 | 2 | Grade 4 |

From Table III above can be seen that the final testing of 12 x-ray images there's still have error prediction, ie the number of four images (33.3%). This can be caused by the weights of the training has not adequately represent the characteristics of each level of the osteoarthritis disease. So when applied to the x-ray images that has not been tested or trained (non-learning), error prediction still can occur. However, if it seen from the results of measurement accuracy percentage at 66.6%, the value is still quite good. And this method can be considered quite successful in determining the disease severity of osteoarthritis. Because the amount of diagnostic accuracy percentage is still higher than the total percentage of failures.

IV. CONCLUSSION

The several of conclusion which can be drawn from this study ie :
1. The use of backpropagation artificial neural network method proved to be as one method to classify or predict the disease severity of osteoarthritis based on color and texture features with accuracy percentage at 66.6%
2. Determination of the learning rate value at 0.3, momentum 0.5, five hidden units, and 132 feature vectors can be stated that the system has been successfully predict the learning data and less successful for non-learning data.